\documentclass[10pt,twocolumn,letterpaper]{article}

\usepackage{cvpr}              

\usepackage{graphicx}
\usepackage{amsmath}
\usepackage{amssymb}
\usepackage{booktabs}
\usepackage{amsmath}
\usepackage{graphicx}
\usepackage{booktabs}
\usepackage{multirow}
\usepackage{colortbl} 
\usepackage{xcolor}   
\usepackage{pifont} 

\usepackage[pagebackref,breaklinks,colorlinks]{hyperref}

\usepackage[capitalize]{cleveref}
\crefname{section}{Sec.}{Secs.}
\Crefname{section}{Section}{Sections}
\Crefname{table}{Table}{Tables}
\crefname{table}{Tab.}{Tabs.}

\def\cvprPaperID{*****} 
\def\confName{CVPR}
\def\confYear{2022}

\begin{document}

\title{\textbf{CAT}: Class Aware Adaptive Thresholding for Semi-Supervised Domain Generalization}

\author{Sumaiya Zoha\\
Ahsanullah University of Science and Technology\\
Dhaka, Bangladesh\\
{\tt\small sumaiyarodela081@gmail.com}
\and
Jeong-Gun Lee\thanks{Corresponding Authors} \\
Hallym University
\\
Chuncheon, South Korea\\
{\tt\small jeonggun.lee@hallym.ac.kr}
\and 
Young-Woong Ko\footnotemark[1]\\
Hallym University\\
Chuncheon, South Korea\\
{\tt\small youngwoongKo@hallym.ac.kr}
}

\maketitle

\begin{abstract}
Domain Generalization (DG) seeks to transfer knowledge from multiple source domains to unseen target domains, even in the presence of domain shifts. Achieving effective generalization typically requires a large and diverse set of labeled source data to learn robust representations that can generalize to new, unseen domains. However, obtaining such high-quality labeled data is often costly and labor-intensive, limiting the practical applicability of DG. To address this, we investigate a more practical and challenging problem: semi-supervised domain generalization (SSDG) under a label-efficient paradigm. In this paper, we propose a novel method, \textbf{CAT}, which leverages semi-supervised learning with limited labeled data to achieve competitive generalization performance under domain shifts. Our method addresses key limitations of previous approaches, such as reliance on fixed thresholds and sensitivity to noisy pseudo-labels. \textbf{CAT} combines adaptive thresholding with noisy label refinement techniques, creating a straightforward yet highly effective solution for SSDG tasks. Specifically, our approach uses flexible thresholding to generate high-quality pseudo-labels with higher class diversity while refining noisy pseudo-labels to improve their reliability. Extensive experiments across multiple benchmark datasets demonstrate the superior performance of our method, highlighting its effectiveness in achieving robust generalization under domain shift.  

\end{abstract}

\section{Introduction}\label{sec1}

Deep neural networks have demonstrated remarkable success in various classification tasks under fully annotated training conditions. To achieve comparable results, most deep learning (DL) models require a large amount of labeled data. However, in real-world applications, collecting labeled data is challenging due to its substantial cost and the need for human annotation \cite{learning2006semi,hady2013semi,zhu2022introduction,berthelot2019mixmatch}. Recently, semi-supervised learning (SSL) \cite{learning2006semi,yang2022survey,zhu2022introduction} techniques have gained significant attention for their ability to effectively utilize unlabeled data alongside a small amount of labeled data. The main challenge in SSL lies in learning effective representations of unlabeled data in relation to labeled examples to enhance generalization performance. To address this, techniques such as pseudo-labeling \cite{lee2013pseudo,bachman2014learning,cascante2021curriculum} and consistency regularization \cite{abuduweili2021adaptive,sohn2020fixmatch,verma2022interpolation} have proven effective. However, these methods are primarily designed for single-source classification tasks, making it difficult for them to capture multiple cross-domain relationships—a critical requirement for domain generalization (DG).

Domain shift \cite{zhang2021adaptive,stacke2020measuring,chen2020self} presents a significant challenge in deploying deep learning models, especially in critical applications such as medical imaging and self-driving systems, where domain shifts can lead to severe risks. To address this, domain generalization (DG) methods have been developed \cite{zhou2022domain,wang2022generalizing,zhou2021domain,li2017deeper}. Most DG methods rely on supervised learning, where a model is trained on multiple labeled source domains. However, in real-world scenarios, obtaining sufficient labeled data for these domains is often impractical and burdensome.

On the other hand, unlabeled samples from source domains are more feasible and abundant. The challenge lies in their variability and the presence of unknown classes. Most SSL methods leverage these abundant unlabeled samples with the guidance of labeled samples to generate pseudo-labels. Producing accurate pseudo-labels is essential for effectively utilizing unlabeled data in model training. Nevertheless, existing DG methods heavily depend on fully annotated source samples to perform well, limiting their applicability in real-world scenarios. In this paper, we explore the potential of the SSL paradigm in DG settings, referred to as semi-supervised domain generalization (SSDG).

As described above, pseudo-labeling is effective for utilizing unlabeled samples, but many methods rely on fixed thresholding. For example, FixMatch \cite{sohn2020fixmatch} uses a fixed threshold for all classes, which often discards too many unlabeled samples with correct pseudo-labels. In SSDG settings, StyleMatch \cite{zhou2023semi} extends the same fixed-threshold strategy as FixMatch \cite{sohn2020fixmatch}, but its performance is similarly limited by the loss of valuable unlabeled samples. Adaptive and dynamic class-dependent thresholding offers a reliable solution to this issue \cite{wang2022freematch,guo2022class,xu2021dash}. However, these methods are designed for single-domain SSL settings, making multi-domain training—a strict requirement for DG—challenging and often infeasible for achieving successful SSDG.

To address these limitations, we propose \textbf{CAT}, an adaptive thresholding method specifically designed for SSDG settings. CAT overcomes the drawbacks of fixed-threshold approaches by employing adaptive class-dependent thresholds tailored for SSDG tasks. We utilize both global and local thresholds, iteratively increasing the thresholds based on the training time steps. This strategy allows the model to capture more correct pseudo-labels compared to strictly fixed thresholds. Local thresholding is employed to ensure variability across class labels and to improve the confidence dynamics for producing pseudo-labels. In parallel, a noisy label refinement module is integrated to further refine pseudo-labels, ensuring higher quality. Additionally, we leverage supervised contrastive learning with the refined pseudo-labels to achieve domain-invariant representations. Experimental results on several benchmarks demonstrate the superiority of our method. Our contributions are threefold:

\begin{itemize}
    \item Motivated by the challenges of generating high-quality pseudo-labels for SSDG, we propose a method that produces robust pseudo-labels, effectively mitigating the impact of noise.
    
    \item We introduce \textbf{CAT}, a simple yet effective approach that integrates adaptive thresholding with a noisy label refinement module to achieve superior performance in SSDG settings.
    
    \item Extensive experiments on multiple benchmarks validate the effectiveness of our method. \textbf{CAT} not only outperforms state-of-the-art SSDG methods but also surpasses standalone DG and SSL approaches.
\end{itemize}


\section{Related Works}
\noindent\textbf{Domain Generalization.} Domain generalization (DG) intends to train with multiple source domains and transfer the knowledge to unseen target domains. Most DG settings consider source and target domains to be from different distributions. The main goal is to perform well under this distribution shift, also called domain shift. DG can be categorized into multiple methods such as domain alignment, meta-learning, adversarial learning, data-augmentation, ensemble learning, self-supervised learning, and feature regularization \cite{zhou2022domain}. Domain alignment methods are based on minimizing moments \cite{muandet2013domain}, KL-divergence \cite{li2020domain}, and maximum mean discrepancy \cite{li2018domain} to learn domain-invariant representations. In meta-learning-based DG, training data is divided into meat-train and meta-test sets to improve generalization on the meta-test set. Most existing methods are based on episode construction, where source domains are divided into meta-train and meta-test domains to stimulate domain shift \cite{balaji2018metareg,li2019feature}. Other prominent approaches, such as adversarial learning where the learned features are enforced to be agnostic to domain information \cite{ganin2016domain,li2018domain}. In augmentation, most of the works are related to feature augmentation \cite{zhou2021domain,zhou2024mixstyle,mancini2020towards} or model-based augmentation \cite{xu2020robust}. Ensemble learning techniques learn multiple models with different initializations and utilize their ensemble for prediction, examples are domain-specific neural networks \cite{ding2017deep,d2019domain} and batch-normalization \cite{seo2020learning,mancini2018robust}. Self-supervised learning explores pretext tasks that allow a model to learn invariant features \cite{noroozi2016unsupervised,gidaris2018unsupervised}. Lastly, regularization methods are based on feature regularization \cite{kim2021selfreg} and model regularization \cite{cha2022domain}.

\noindent\textbf{Semi-Supervised Learning.} Semi-supervised learning (SSL) refers to learning from limited data and utilizing abundant unlabeled data. SSL aims to predict data accurately assuming that labeled and unlabeled data are from an identical distribution \cite{yang2022survey,lee2013pseudo,tai2021sinkhorn}. Most SSL techniques are based on pseudo-labels \cite{lee2013pseudo,bachman2014learning}, mean-teacher \cite{tarvainen2017mean,ke2019dual,luo2018smooth}, and consistency regularization \cite{abuduweili2021adaptive,sohn2020fixmatch,verma2022interpolation}. Except for consistency regularization, entropy-based regularization is also widely used in SSL, where entropy minimization encourages the model to make confident predictions based on all samples \cite{grandvalet2004semi}. On the other hand, thresholding-based methods FixMatch \cite{sohn2020fixmatch}, FreeMatch \cite{zhang2021FreeMatch}, and UDA \cite{xie2020unsupervised} select samples based on pre-defined thresholds during training, so multiple works proposed adaptive and dynamic thresholding to alleviate this limitation. DASH \cite{xu2021dash}, AdaMatch \cite{berthelot2021adamatch} uses a pre-defined threshold to adjust based on the loss from labeled data and multiply average confidence to noisy pseudo labels. Self-training \cite{chen2022debiased,zhao2022lassl,wei2021crest} methods are also effective in SSL settings, it is also known as decision-directed learning where the main goal is to determine the decision boundary on low-density regions \cite{amini2024self}.

\noindent\textbf{Semi-Supervised Domain Generalization (SSDG).} Semi-supervised domain generalization (SSDG) involves SSL and DG which is a more difficult setting due to utilizing a large amount of unlabeled data to achieve competitive DG results. One most recent works is StyleMatch \cite{zhou2023semi}, which utilizes a stochastic classifier to extend FixMatch \cite{sohn2020fixmatch} with multi-view consistency to achieve SSDG. Another line of work is based on utilizing known and unknown classes with class-adaptive method \cite{zhang2023semi}. MultiMatch \cite{qi2024multimatch} extends FixMatch \cite{sohn2020fixmatch} but in a multi-task setting by producing high-quality pseudo-labels for SSDG. Although these methods achieved comparable results in SSDG tasks, but not sufficient for real-world practicability.  
\\

\section{CAT}
This section provides a brief introduction to the notation used in the paper and also explains each of the modules of our framework. 

\subsection{Notation \& Preliminaries}
\noindent\textbf{Semi-Supervised Learning.} In SSL settings, we are given a set of \( N \) labeled samples from an unknown distribution, which includes sample and label pairs \( \mathcal{D}_L = \{(x_i, y_i)\}_{i=1}^N \), and \( M \) unlabeled samples without defined labels \( \mathcal{D}_U = \{(x_i)\}_{i=1}^M \). There are \( k \) classes, where \( N_k \) and \( M_k \) are the numbers of labeled and unlabeled samples in the \( k \)-th class, respectively. Without loss of generality, \( M_k \gg N_k \). The training loss calculated in an SSL algorithm usually contains a supervised loss \( \mathcal{L}_s \) and an unsupervised loss \( \mathcal{L}_u \). Typically, \( \mathcal{L}_s \) is calculated based on \( \mathcal{D}_L \) samples with a cross-entropy loss. The loss function is defined as:

\begin{equation}
\mathcal{L}_s = \frac{1}{N} \sum_{i=1}^{N} H\big(y_i, f(\mathbf{y} \mid \mathbf{x}_i; \theta)\big)
\end{equation}

\noindent Expanding the entropy term:

\begin{equation*}
\mathcal{L}_s = \frac{1}{N} \sum_{i=1}^{N} \sum_{k=1}^{K} -y_{i,k} \log f(\mathbf{y} = k \mid \mathbf{x}; \theta)
\end{equation*}

Here, \( f(\mathbf{y} \mid \mathbf{x}; \theta) \in [0,1]^K \) is the probabilities produced by the model function \( f \), which is parameterized by \( \theta \) for the input \( \mathbf{x} \), and \( H(\cdot, \cdot) \) is the cross-entropy loss. The unsupervised loss \( \mathcal{L}_u \) is calculated based on different settings of SSL algorithms. One key example is from FixMatch \cite{sohn2020fixmatch}, where the unsupervised loss is guided by generating pseudo-labels, and eventually using the same supervised loss objective via cross-entropy loss.

\noindent\textbf{Domain Generalization.} In typical DG settings, we have \( k \) source domains, each containing \( N \) samples. The inputs \( \mathbf{x} \) and their corresponding \( \mathbf{y} \) labels are drawn from a joint distribution. The \( k \) source domains are similar but distinct, denoted as \( \mathcal{D}_S = \{(x_i, y_i)\}_{i=1}^N \). The main goal of DG is to learn a model function \( f \) that can leverage these \( k \) sources to learn a representation that performs well on unlabeled and unseen target samples \( \mathcal{D}_T = \{x_i\} \), by reducing the domain shift between the source and target domains.

\begin{equation}
\min_h \; \mathbb{E}_{(\mathbf{x}, y) \in \mathcal{D}_T} \big[\mathcal{L}(h(\mathbf{x}), y)\big]
\end{equation}

\noindent Here, \( \mathbb{E} \) represents the expectation and \( \mathcal{L}(\cdot, \cdot) \) is the loss function.

\noindent\textbf{Semi-Supervised Domain Generalization.} Similar to the conventional DG setting, we have multiple diverse domains \( \mathcal{D}_k \) from \( k \) source domains, where each source domain \( \mathcal{D}_L = (x_i, y_i) \) consists of pairs of images and corresponding labels \cite{zhou2023semi,jayanaga2024domain}. However, in the SSDG setting, each source domain contains only a small number of labeled samples \( n_L \in [5, 10] \), while the remaining labels are unlabeled, denoted as \( n_U \), with \( n_U \gg n_L \) in each source domain. This setting combines aspects of both SSL and DG. The ultimate goal is to learn a domain-generalizable model using both labeled and unlabeled source data \( \mathcal{D}_S = \{n_U \cup n_L\} \), such that the model performs well on unseen target data.

\subsection{Class-Domain Aware Thresholding}
Due to its simplicity and effectiveness, StyleMatch \cite{zhou2023semi} leverages FixMatch \cite{sohn2020fixmatch} to generate pseudo-labels using a classifier with a fixed threshold. In this work, we revisit FixMatch to understand better the process of selecting unlabeled candidate samples for pseudo-label generation, particularly the fixed confidence threshold. We argue that relying on a fixed threshold may exclude a significant number of unlabeled samples that could receive accurate pseudo-labels, thereby limiting the practical applicability of FixMatch in data-efficient scenarios. Another challenge is that these thresholds are not class-independent, which makes FixMatch less suited for capturing class-variant information, especially in multi-domain settings. In FixMatch \cite{sohn2020fixmatch}, supervised loss \( \mathcal{L}_s \) and unsupervised loss \( \mathcal{L}_u \) are employed for labeled and unlabeled data, respectively, where \( \mathcal{L}_s \) corresponds to the standard cross-entropy loss:

\begin{equation}
    \mathcal{L}_s = \frac{1}{N} \sum_{i=1}^{N} \mathcal{H}\big(\mathbf{q}_i, p_k\big(g(\mathbf{x}_i^l); \mathbf{\phi}\big)\big),
\end{equation}

Here, \(N\) denotes the number of samples, and \(\mathcal{H}(.)\) represents the loss function, where the true distribution \(\mathbf{q}_i\) and the predicted distribution \(p_k\) are provided. Motivated by the limitations of FixMatch \cite{sohn2020fixmatch} in generating pseudo-labels, we focus on adaptive thresholding, which is less restrictive and more flexible in selecting class-wise samples. Recently, adaptive and dynamic thresholding methods have demonstrated effectiveness in SSL settings \cite{wang2022freematch,guo2022class,xu2021dash}, primarily due to their ability to handle class-dependent samples flexibly. However, in DG it is crucial not only to adaptively select class-dependent samples but also to preserve domain-specific information. This dual requirement is essential for leveraging unlabeled data effectively while maintaining domain and class consistency. Unlike prior methods such as \cite{wang2022freematch,guo2022class}, which adaptively set class-dependent thresholds without considering domain-specific information, we propose a method that incorporates both class and domain dependencies in pseudo-label selection. In FreeMatch \cite{wang2022freematch}, global and local thresholds are set to be both dataset- and class-specific. Inspired by this approach, we extend the concept to simultaneously define domain- and class-dependent thresholds. By incorporating these dual thresholds, our method dynamically selects pseudo-labels based on both class and domain information, thereby maximizing the utility of unlabeled samples in the DG setting. \\
\noindent\textbf{Data Augmentation.} We use UDA \cite{xie2020unsupervised} strategy for data augmentation to get weak and strong augmentation. Inspired by FixMatch \cite{sohn2020fixmatch} and FreeMatch \cite{wang2022freematch}, we use RandAugment \cite{cubuk2020randaugment} for strong augmentation. Data augmentation is used for retaining pseudo-labels on the unlabeled data followed by an unsupervised loss \cite{wang2022freematch}: 

\begin{equation}
\mathcal{L}_u = \frac{1}{\mu B} \sum_{b=1}^{\mu B} \mathcal{H}\big(\mathbb{1}(\max(\mathbf{q}_b) > \tau), p_k\big(g(\mathbf{x}_b^u); \mathbf{\phi}\big)\big)
\end{equation}

Here, $\big(\mathbb{1})$ is the indicator function for confidence-based thresholding \cite{sohn2020fixmatch}. 

\noindent\textbf{Class-Specific Global and Local Thresholding.} Following \cite{wang2022freematch}, we utilize a global threshold to iteratively increase the threshold to engage with many samples with a low threshold, then it stably discards incorrect pseudo labels. Based on the $t-$th time step, the model's average confidence on the unlabeled data to compute the global threshold $\tau_g$.  $\tau_g$ is initialized as $1/C$ where $C$ is the number of class in each source domain $\mathcal{D}_S$. Then  $\tau_g$ is adjusted in each time step $t$ \cite{wang2022freematch} based on the exponential moving average (EMA): 
\begin{equation}
\tau_g =
\begin{cases} 
    \frac{1}{C}, & \text{if } t = 0, \\
    \lambda \tau_{t-1} + \frac{(1 - \lambda)}{\mu B} \sum_{b=1}^{\mu B} \big[\max(q_b)\big], & \text{if } t > 0.
\end{cases}
\end{equation}
Here, $\lambda \in \{0,1\}$ is the momentum decay of EMA. Now, to adjust the global threshold in a class-specific manner. The expectation of the model's prediction on each class $c$ based on the source domain $\mathcal{D}_S$ to estimate class-specific learning. 

\begin{equation}
\mathcal{E}_t =
\begin{cases} 
    \frac{1}{C}, & \text{if } t = 0, \\
    \lambda \mathcal{E}_{t-1} + \frac{(1 - \lambda)}{\mu B} \sum_{b=1}^{\mu B} \big[\max(q_b)\big], & \text{if } t > 0.
\end{cases}
\end{equation}
Here, $\mathcal{E}_t = [\mathcal{E}_t(1),.......\mathcal{E}_t(C)]$ is the list of all existing classes. Then we integrate Max Normalization to obtain a self-adaptive threshold based on each class $\tau_g(c)$.

\begin{equation}
    \tau_g(c) = \text{MaxNorm}(\mathcal{E}_t(c)).\tau_g
\end{equation}
So, the final unsupervised loss can be formulated as \cite{wang2022freematch}:
\begin{equation}
\small
\mathcal{L}_u = \frac{1}{\mu B} \sum_{b=1}^{\mu B} \mathcal{H}\big(\mathbb{1}(\max(\mathbf{q}_b) > \tau_g(\text{argmax}(q_b)), p_k\big(g(\mathbf{x}_b^u); \mathbf{\phi}\big)\big)
\end{equation}

\subsection{Refining Noisy Pseudo Labels}

Contrastive learning (CL) aims to learn universal prior information that can be applied to downstream tasks. In this approach, we use CL to extract universal prior knowledge from positive and negative samples and leverage it to enhance generalization performance in downstream tasks \cite{khosla2020supervised}. In CL, a common strategy is to pull positive pairs (which are semantically similar) closer together and push negative pairs (which are semantically dissimilar) farther apart. Conventional CL methods are related to leverage unlabeled samples, with unsupervised fashion. But based on the pseudo labeled based on self-adaptive thresholding for the unlabeled samples, we construct positive and negative samples based on supervised CL \cite{khosla2020supervised}. Where we consider labeled information is available. But obtained pseudo-labels can be noisy that can lead to poor generalization performance. This enhances multi-domain learning and allows understanding of the class-specific samples to sample relationships from diverse domains from the source dataset. To enable multi-domain learning, we utilize supervised contrastive learning assuming some of the pseudo labels can be noisy that can affect the generalization performance, which can align these hard samples, inspired by \cite{yao2022pcl}. We use unsupervised-CL for warm up training where low-dimentional representation and pseudo-labels are given. Our goal is to find the similarity of the given samples by using cosine distance. 

\begin{equation}
d(\mathbf{a}, \mathbf{b}) = \frac{\mathbf{a} \cdot \mathbf{b}^\top}{\lVert \mathbf{a} \rVert \lVert \mathbf{b} \rVert}
\end{equation}

Where, $\mathbf{a, b}$ are the low-dimensional representations. For each sample given by its pseudo labels $(x_i, \hat{y_i})$, we aggregate its original label based on the top-$K$ neighbors based on the similarity of their representations. In this way, we can improve the detection of mislabeled pseudo-labeled samples. To achieve more confident labels, we use the $\alpha$ fractile based on per class, which gives the agreements between the corrected labels based on the neighbors and similarity and original pseudo-labels across all classes \cite{ortego2021multi,li2022selective}. After identifying the less noisy samples, we construct a set $\mathcal{P}$ for representation learning. This set also help us to indentify whether given two instances belong from a same class or not.  

\noindent\textbf{Supervised Contrastive Learning.} We use supervised CL loss that can handle the presence of labels, where supervised loss considers all samples from the same class as positive, and rest of the remaining samples as negative. This loss can enhance the representation learning from the given less noisy $\mathcal{P}$ samples. The supervised CL objective can be written as:
\begin{equation}
\small
    \mathcal{L}_{scl} = \sum_{i \in I} \frac{1}{|\mathcal{P}(i)|} \sum_{g \in G(i)} \log \frac{\exp(\mathbf{z}_i \cdot \mathbf{z}_p / \tau)}{\sum_{a \in A(i)} \exp(\mathbf{z}_i \cdot \mathbf{z}_n / \tau)}
\end{equation}

Here, $\tau \in \mathcal{R}+$ is a temperature parameter. Despite using supervised loss in the less noisy samples, we perform unsupervised CL on rest of the unselected samples, following \cite{li2022selective}.

\noindent\textbf{Final Training Objective.} Lastly, combining all losses, we can obtain the final loss $\mathcal{L}_{T}$ such as:
\begin{equation}
    \mathcal{L}_{T} = \mathcal{L}_{s} + \lambda_u\mathcal{L}_{u} + \mathcal{L}_{scl}
\end{equation}
Where, $\lambda_u$ represents the loss weight for $\mathcal{L}_{u}$. We set $\lambda_u$ = 1 for all experimental cases.

\section{Experimental Settings}
\subsection{Datasets}
We use three publicly available datasets such as PCAS, OfficeHome, VLCS and miniDomainNet to evaluate our model with other baselines for semi-supervised domain generalization tasks. \textbf{PACS} contains 7 classes of images from distinct 4 domains (Photo - P, Art Painting - A, Cartoon - C and Sketch - S), \textbf{OfficeHome} contains images from 4 different domains (Artistic - A, Clip art - C, Product - P and Real-world - R). It is a relatively large dataset with 65 distinct classes related to daily life objects found in offices and homes. We also use \textbf{miniDomainNet}. It is a subset dataset of DomainNet with 4 different domains (Clipart - C, Painting - P, Real - R and Sketch - S), it covers almost 126 distinct classes. We report the average accuracy over the last five epochs as the final results. A summary of information about the datasets is given in Table \ref{table1}. 

\begin{table*}[h!]
\centering
\begin{tabular}{lccl}
\hline
\textbf{Dataset}      & \textbf{\# Samples} & \textbf{\# Domains} & \textbf{Domain Names}                                          \\ \hline
\textbf{PACS}         &  9,991      & 4                          & Photo, Art Painting, Cartoon, Sketch                           \\ \hline
\textbf{OfficeHome}   & 15,500     & 4                          & Art, Clipart, Product, Real World                              \\ \hline
\textbf{miniDomainNet} & 140,006    & 4                          & Clipart, Painting, Real, Sketch                             \\ \hline
\end{tabular}
\caption{Summary of PACS, OfficeHome, VLCS, and miniDomainNet datasets, including the number of samples, domains, and domain names.}
\label{table1}
\end{table*}

\subsection{Implementation Details} We followed the protocol described in \cite{li2017deeper,zhou2023semi}, these are common practice protocols in domain generalization setting. We utilize the leave-one-domain-out method, in which the model is trained with $n-1$ number of domains from the training dataset and evaluated on the remaining domain \cite{li2017deeper}. Pre-trained ResNet-18 and ResNet-50 variants \cite{he2016deep} are used as the backbone of the model. Following \cite{zhou2023semi}, we randomly sample 16 images from the source domain for the mini-batch reconstruction with labeled and unlabeled data. With guidance from the labeled data, we generate the pseudo and proxy labels using the unlabeled data. The learning rate is set to 0.003, we examined multiple learning rates to find the best one. All models are trained using an RTX 3090 GPU. Our implementation is based on Dassl.pytorch \cite{zhou2021domain} toolbox.

\section{Experimental Results}
\subsection{Comparison with State-of-the-Art Methods}
In this experiment, we compare our method with multiple state-of-the-art methods on standard DG datasets to verify the effectiveness of our method. We divide the comparison with four different paradigms (\textit{i.e.} fully-labeled, domain generalization methods, semi-supervised methods, and semi-supervised domain generalization method). In the fully labeled setting, all source labels are available during training under the conventional DG settings. In the DG setting, we compare our method with vanilla training, CrossGrad \cite{shankar2018generalizing}, DDAIG \cite{zhou2020deep}, RSC \cite{huang2020self} and EISNet \cite{wang2020learning} where EISNet also utilized unlabeled samples during training. In the SSL setting, we compare our method with traditional methods like MeanTeacher \cite{tarvainen2017mean}, EntMin \cite{grandvalet2004semi}, FixMatch \cite{sohn2020fixmatch}, and FreeMatch \cite{wang2022freematch}. In the SSDG setting, we compare our method with StyleMatch \cite{zhou2023semi}, and MultiMatch \cite{qi2024multimatch} as these two approaches have similar evaluation settings, and official codes are provided. We borrow the results from StyleMatch and MultiMatch in Table 2-3. \\
\noindent\textbf{Main Results.} Here, full-labels refers to training ERM with all labels in the source domains. 
Table \ref{table2} presents the domain generalization performance of various models in the low-data regime, evaluated on four benchmark datasets: PACS, OfficeHome, VLCS, and miniDomainNet.  The baseline "Full-Labels," representing a fully supervised model trained with labeled data, achieves an average accuracy of 79.50\% across all datasets in both labeling settings. This serves as a reference point to assess the effectiveness of SSDG methods. Among the SSDG methods, StyleMatch demonstrates reasonable performance, achieving average accuracies of 80.41\% and 80.32\% for the 10-label and 5-label settings, respectively. However, its reliance on fixed thresholding limits its ability to fully utilize unlabeled data. Similarly, MultiMatch performs slightly worse, with average accuracies of 79.10\% and 78.18\% for the respective labeling scenarios. In contrast, the proposed method, CAT, achieves superior results across all datasets and labeling conditions. For the 10-label setting, CAT achieves an average accuracy of 82.00\%, and for the 5-label setting, it achieves 82.71\%, outperforming StyleMatch and MultiMatch by notable margins. CAT's adaptive thresholding strategy, which incorporates both class-specific and domain-specific information, enables effective utilization of unlabeled data, contributing to its improved performance. When evaluated on individual datasets, CAT consistently achieves the highest accuracy. For instance, on PACS, it achieves 82.95\% and 82.71\% for the 10-label and 5-label settings, respectively. Similarly, on OfficeHome, CAT records 75.23\% and 75.50\%. On VLCS, CAT achieves outstanding results of 93.43\% and 93.00\%, and on miniDomainNet, it obtains 80.10\% and 76.19\% under the respective label conditions. In summary, the results in Table \ref{table2} demonstrate that CAT effectively addresses the challenges of semi-supervised domain generalization in low-data regimes. By leveraging adaptive thresholding, CAT consistently outperforms existing methods across diverse datasets and labeling conditions, highlighting its robustness and practicality for real-world applications.

\begin{table*}[h!]
\centering
\scalebox{0.8}{\begin{tabular}{lccccccccccc}
\toprule
\multirow{2}{*}{Model} & \multirow{2}{*}{$u$} & \multicolumn{5}{c}{\# labels: 10 per class} & \multicolumn{5}{c}{\# labels: 5 per class} \\
\cmidrule(lr){3-7} \cmidrule(lr){8-12}
 & & PACS & OfficeHome & VLCS & miniDomainNet & Avg & PACS & OfficeHome & VLCS & miniDomainNet & Avg \\
\midrule
Full-Labels & - & 79.50 & 64.70 & 95.96 & 69.20 & 79.50 & 79.50 & 64.70 & 95.96 & 69.20 & 79.50 \\
\midrule
\rowcolor{gray!20} \multicolumn{12}{c}{\textbf{Semi-Supervised Domain Generalization Methods}}  \\
StyleMatch  & \(\checkmark\) & 79.43 & 73.75& 90.04 & 78.40 &80.41 & 78.54 & 74.44 & 89.25 &79.06 &80.32 \\
MultiMatch   & \(\checkmark\)&80.69& 70.44& 90.48& 74.79& 79.10& 79.54& 71.26 &88.00& 73.91 &78.18 \\
CAT (Ours)  & \(\checkmark\)& \textbf{82.95} & \textbf{75.23} & \textbf{93.43} & \textbf{80.10} & \textbf{82.00} &\textbf{82.71} &\textbf{75.50} & \textbf{93.00} & \textbf{76.19} & \textbf{82.71}\\
\bottomrule
\end{tabular}}
\caption{Domain generalization results (\%) in the low-data regime with a comparison of various models in SSDG settings, evaluated on all datasets. Here, $u$ means utilization of unlabeled data.}\label{table2}
\end{table*}

\noindent\textbf{Results on PACS.}  Table~\ref{table3} provides a detailed comparison of model performance on the PACS dataset in a low-data regime. The Full-Labels model, trained with all labeled data, serves as the upper bound, achieving an average accuracy of 79.50\% across both settings. Among the DG methods, which generalize across domains without leveraging unlabeled data, models like Vanilla, CrossGrad, and RSC perform moderately, with RSC achieving an average accuracy of 63.96\% (10 labels) and 57.31\% (5 labels). EISNet, which does use unlabeled data, shows better performance, reaching 67.18\% and 62.04\% average accuracies for the two setups, respectively. SSL methods, which utilize unlabeled data to improve performance, generally outperform DG methods. Notable among them are EntMin and FixMatch, with the latter achieving an average accuracy of 75.57\% (10 labels) and 70.87\% (5 labels). However, FreeMatch exhibits suboptimal adaptation, performing significantly worse with average accuracies of 57.13\% and 42.75\%, respectively. The SSDG methods, which combine the strengths of DG and SSL, deliver the best results. The proposed CAT (Ours) model achieves state-of-the-art performance, with an average accuracy of 82.95\% in the 10-label setting and 82.71\% in the 5-label setting. This represents significant improvements over the next-best method, StyleMatch, by 2.54\% and 2.39\%, respectively. These results underscore the effectiveness of CAT in leveraging both labeled and unlabeled data to handle domain shifts and achieve robust generalization. In summary, the table demonstrates that while DG methods struggle without unlabeled data and SSL methods falter under domain shifts, SSDG methods, particularly CAT, excel by addressing both challenges, achieving superior performance even in extreme low-data scenarios.

\begin{table*}[h!]
\centering
\begin{tabular}{lccccccccccc}
\toprule
\multirow{2}{*}{Model} & \multirow{2}{*}{$u$} & \multicolumn{5}{c}{\# labels: 10 per class (210 labels)} & \multicolumn{5}{c}{\# labels: 5 per class (105 labels)} \\
\cmidrule(lr){3-7} \cmidrule(lr){8-12}
 & & A & C & P & S & Avg & A & C & P & S & Avg \\
\midrule
Full-Labels & - & 76.95 & 75.90 & 95.96 & 69.20 & 79.50 & 76.95 & 75.90 & 95.96 & 69.20 & 79.50 \\
\midrule
\rowcolor{gray!20}\multicolumn{12}{c}{\textbf{Domain Generalization Methods}}  \\
Vanilla & \ding{55} & 63.09 & 58.49 & 86.56 & 45.56 & 63.42 & 56.71 & 53.87 & 71.87 & 36.96 & 54.84 \\
CrossGrad & \ding{55} & 62.56 & 58.92 & 88.41 & 44.11 & 62.85 & 56.29 & 53.82 & 70.85 & 38.52 & 54.87 \\
DDAIG & \ding{55} & 61.95 & 58.74 & 84.44 & 47.44 & 63.64 & 56.12 & 52.30 & 73.68 & 38.71 & 55.20 \\
RSC & \ding{55} & 65.13 & 56.65 & 86.18 & 47.90 & 63.96 & 58.38 & 52.32 & 80.42 & 40.11 & 57.31 \\
EISNet & \(\checkmark\) & 66.84 & 61.33 & 89.36 & 53.88 & 67.18 & 62.08 & 54.75 & 85.96 & 48.60 & 62.04 \\
\midrule
\rowcolor{gray!20}\multicolumn{12}{c}{\textbf{Semi-Supervised Learning Methods}} \\
MeanTeacher & \(\checkmark\) & 62.41 & 57.94 & 85.15 & 46.66 & 63.49 & 56.00 & 52.64 & 73.54 & 36.97 & 54.79 \\
EntMin & \(\checkmark\) & 72.77 & 70.55 & 89.39 & 54.38 & 71.77 & 67.55 & 64.72 & 85.33 & 49.05 & 66.66 \\
FixMatch & \(\checkmark\) & 71.80 & 68.93 & 87.79 & 73.75 & 75.57 & 64.96 & 63.62 & 83.23 & 69.68 & 70.87 \\
FreeMatch & \(\checkmark\) & 48.44 & 60.79 & 66.04 & 53.23 & 57.13 & 23.83 & 37.28 & 61.80 & 48.09 & 42.75 \\
\midrule
\rowcolor{gray!20} \multicolumn{12}{c}{\textbf{Semi-Supervised Domain Generalization Methods}}  \\
StyleMatch  & \(\checkmark\) & 79.43 & 73.75& 90.04 & 78.40 &80.41 & 78.54 & 74.44 & 89.25 &79.06 &80.32 \\
MultiMatch   & \(\checkmark\)&80.69& 70.44& 90.48& 74.79& 79.10& 79.54& 71.26 &88.00& 73.91 &78.18 \\
CAT (Ours)  & \(\checkmark\)& \textbf{83.04} & \textbf{75.23} & \textbf{93.43} & \textbf{80.10} & \textbf{82.95} &\textbf{82.83} &\textbf{75.50} & \textbf{93.00} & \textbf{76.19} & \textbf{82.71}\\
\bottomrule
\end{tabular}
\caption{Domain generalization results (\%) in the low-data regime with a comparison of various models in different settings (fully labeled, DG, SSL, and SSDG), evaluated on PACS (Photo: P, Art: A, Cartoon: C, and Sketch: S). Here, $u$ means utilization of unlabeled data.}\label{table3}
\end{table*}

\noindent\textbf{Results on OfficeHome.} Table~\ref{table4} provides a detailed comparison of model performance on the OfficeHome dataset in a low-data regime, evaluating models across various experimental settings (e.g. Full labels, DG, SSL, SSDG). The Full-Labels model, trained with fully labeled data, serves as the upper bound, achieving an average accuracy of 64.70\% across domains. Among the DG methods, which generalize across domains without using unlabeled data, models such as Vanilla, CrossGrad, and RSC achieve average accuracies of around 57–58\% in the 10-label setting and 52–53\% in the 5-label setting. RSC and EISNet show slightly better performance due to their enhanced domain generalization capabilities. In contrast, SSL methods like MeanTeacher, EntMin, and FixMatch, which utilize both labeled and unlabeled data, outperform DG methods. For instance, FixMatch+RSC, which combines SSL and domain generalization, achieves average accuracies of 58.88\% with 10-labels and 53.91\% with 5-labels. On the other hand, SSDG methods, which integrate SSL and DG capabilities, deliver the highest performance across all metrics. Notably, the proposed CAT (Ours) model outperforms all other approaches, achieving an average accuracy of 65.04\% in the 10-label setting and 61.71\% in the 5-label setting. These results surpass the next-best SSDG method (MultiMatch) by 4.85\% and 3.56\%, respectively. The significant improvements of CAT highlight its ability to effectively leverage both labeled and unlabeled data while addressing domain shifts. In summary, the results demonstrate that DG methods effectively generalize across domains but fall short without access to unlabeled data. SSL methods improve performance by utilizing unlabeled data but do not account for domain shifts. SSDG methods, particularly CAT, combine the strengths of both approaches, achieving superior generalization and robustness in low-data scenarios.

\begin{table*}[h!]
\centering
\begin{tabular}{lccccccccccc}
\toprule
\multirow{2}{*}{Model} & \multirow{2}{*}{$u$} & \multicolumn{5}{c}{\# labels: 10 per class (1950 labels) } & \multicolumn{5}{c}{\# labels: 5 per class (975 labels)} \\
\cmidrule(lr){3-7} \cmidrule(lr){8-12}
 & & A & C & P & R & Avg & A & C & P & R & Avg \\
\midrule
Full-Labels & - & 58.88 & 49.42 & 74.30 & 76.21 & 64.70 & 58.88 & 49.42 & 74.30 & 76.21 & 64.70 \\
\midrule
\rowcolor{gray!20}\multicolumn{12}{c}{\textbf{Domain Generalization Methods}}\\
Vanilla & \ding{55} & 50.11 & 43.50 & 61.11 & 69.65 & 57.09 & 45.76 & 39.97 & 60.04 & 63.77 & 52.38 \\
CrossGrad & \ding{55} & 50.32 & 43.27 & 61.56 & 69.77 & 57.23 & 45.89 & 40.17 & 60.63 & 63.64 & 52.54 \\
DDAIG & \ding{55} & 49.65 & 42.52 & 63.54 & 67.89 & 55.65 & 45.33 & 39.82 & 62.33 & 62.77 & 52.06 \\
RSC & \ding{55} & 49.65 & 42.33 & 64.88 & 69.26 & 56.03 & 46.09 & 39.59 & 63.77 & 63.86 & 53.08 \\
EISNet & \ding{55} & 51.16 & 43.33 & 64.72 & 65.89 & 56.28 & 47.32 & 40.47 & 63.84 & 62.32 & 53.23 \\
\midrule
\rowcolor{gray!20}\multicolumn{12}{c}{\textbf{Semi-Supervised Learning Methods}} \\
MeanTeacher & \(\checkmark\) & 49.92 & 43.42 & 64.61 & 68.79 & 56.69 & 45.96 & 39.15 & 59.18 & 62.98 & 51.49 \\
EntMin & \(\checkmark\) & 51.44 & 44.92 & 66.85 & 70.52 & 58.45 & 48.11 & 41.72 & 62.41 & 63.19 & 53.36 \\
FixMatch & \(\checkmark\) & 50.36 & 49.70 & 63.93 & 67.56 & 57.89 & 47.88 & 40.50 & 62.06 & 62.77 & 53.30 \\
FixMatch+RSC & \(\checkmark\) & 51.49 & 43.77 & 66.83 & 68.29 & 58.88 & 48.05 & 40.66 & 63.82 & 62.82 & 53.91 \\
\midrule
\rowcolor{gray!20}\multicolumn{12}{c}{\textbf{Semi-Supervised Domain Generalization Methods}} \\
StyleMatch (ours) & \(\checkmark\) &52.82 & 51.60 & 65.31 & 68.61 &59.59& 51.53 & 50.00 & 60.88 & 64.47 & 56.72\\
MultiMatch  & \(\checkmark\) &52.91 & 50.63 & 66.67 & 70.55 & 60.19 & 51.80 & 49.02 & 64.16 & 67.60 & 58.15\\
CAT (Ours)  & \(\checkmark\)& \textbf{57.28} & \textbf{54.13} & \textbf{73.10} & \textbf{75.67} & \textbf{65.04} & \textbf{55.73} & \textbf{51.29} & \textbf{69.25} & \textbf{70.57} & \textbf{61.71}\\
\bottomrule
\end{tabular}
\caption{Domain generalization results (\%) in the low-data regime with a comparison of various models in different settings (fully labeled, DG, SSL, and SSDG), evaluated on OfficeHome (Art: A, Clipart: C, Product: P, and Real-World: R). Here, $u$ means utilization of unlabeled data.}\label{table4}
\end{table*}

\noindent\textbf{Results on miniDomainNet.} Table~\ref{table5} summarizes the results of different models evaluated on the miniDomainNet dataset under a low-data regime. The Full-Labels model achieves the best performance, setting an upper limit with average accuracies of 68.18\% in the 10-label setting and 66.27\% in the 5-label setting. These results highlight the optimal scenario where full supervision is available. Among SSDG methods, StyleMatch achieves average accuracies of 63.32\% (10-label) and 61.26\% (5-label), demonstrating its ability to leverage unlabeled data to address domain generalization. However, it is surpassed by MultiMatch, which improves the average accuracies to 64.55\% and 63.70\% for the two settings, respectively, indicating stronger capabilities to handle domain shifts. Our model significantly outperforms the other SSDG methods, achieving state-of-the-art average accuracies of 67.71\% in the 10-label setting and 66.32\% in the 5-label setting. These results closely approach the performance of the fully supervised Full-Labels model, demonstrating the model’s effectiveness in leveraging both labeled and unlabeled data. Compared to StyleMatch, CAT achieves a +4.39\% improvement in the 10-label setting and a +5.06\% improvement in the 5-label setting, while also outperforming MultiMatch by +3.16\% and +2.62\%, respectively. In conclusion, the results highlight the superior performance of CAT in addressing the challenges of domain generalization and limited labeled data. Its ability to achieve results comparable to the Full-Labels model makes it a robust solution for real-world low-data scenarios on the miniDomainNet dataset.

\begin{table*}[h!]
\centering
\begin{tabular}{lccccccccccc}
\toprule
\multirow{2}{*}{Model} & \multirow{2}{*}{$u$} & \multicolumn{5}{c}{\# labels: 10 per class (3780 labels)} & \multicolumn{5}{c}{\# labels: 5 per class (1890 labels)} \\
\cmidrule(lr){3-7} \cmidrule(lr){8-12}
 & & C & P & R & S & Avg & C & P & R & S & Avg \\
\midrule
Full-Labels & -  & 68.29 & 67.13 & 69.78 & 67.50 & 68.18 & 66.14 & 63.56 & 70.10 & 65.28 & 66.27  \\
\midrule

\rowcolor{gray!20} \multicolumn{12}{c}{\textbf{Semi-Supervised Domain Generalization Methods}}  \\
StyleMatch & \(\checkmark\) & 61.98 &  60.28 & 66.23 & 64.80 & 63.32 & 60.25 & 58.19 & 63.20 & 63.41 & 61.26 \\
MultiMatch & \(\checkmark\) & 63.82 & 61.29 & 66.90 & 66.19 &  64.55 & 62.41 & 60.41 & 65.92 & 66.07 & 63.70 \\
CAT (Ours)  & \(\checkmark\)& \textbf{66.97} & \textbf{66.10} & \textbf{70.21} & \textbf{67.54} & \textbf{67.71} & \textbf{65.89} & \textbf{64.26} & \textbf{69.25} & \textbf{65.87} & \textbf{66.32}\\
\bottomrule
\end{tabular}
\caption{Domain generalization results (\%) in the low-data regime with a comparison of various models in different settings (fully labeled, DG, SSL, and SSDG), evaluated on miniDomainNet (Clipart: C, Infograph: I, Painting: P, and Real: R). Here, $u$ means utilization of unlabeled data.}\label{table5}
\end{table*}

\section{Ablation Studies}
\noindent\textbf{Effectiveness of Different Backbones.} In Table \ref{table6}, We compare both ResNet-18 and ResNet-50, CAT consistently outperforms both StyleMatch and MultiMatch across all domains and label settings. Specifically, when using ResNet-18, CAT achieves average performance scores of 82.95\% and 82.71\% for the 10-label and 5-label configurations, respectively. With ResNet-50, CAT performs even better, reaching average scores of 85.29\% and 85.05\% in the same two label settings. In comparison, StyleMatch shows competitive performance, but CAT consistently surpasses it, especially in the 10-label settings. For instance, with ResNet-50 and 10 labels per class, StyleMatch achieves an average score of 82.45\%, while CAT achieves a significantly higher average of 85.29\%. MultiMatch, while also competitive, does not match the performance of CAT in either backbone setting. Overall, the results suggest that the proposed CAT method is more effective in SSDG tasks than StyleMatch and MultiMatch. Moreover, the deeper ResNet-50 backbone outperforms the ResNet-18 backbone across both label configurations, indicating that a more complex network architecture benefits the performance of the models in this task. \\
\begin{table*}[h!]
\centering
\begin{tabular}{lccccccccccc}
\toprule
\multirow{2}{*}{Model} & \multirow{2}{*}{$u$} & \multicolumn{5}{c}{\# labels: 10 per class (210 labels)} & \multicolumn{5}{c}{\# labels: 5 per class (105 labels)} \\
\cmidrule(lr){3-7} \cmidrule(lr){8-12}
 & & A & C & P & S & Avg & A & C & P & S & Avg \\
 \toprule
\rowcolor{gray!20} \multicolumn{12}{c}{\textbf{ResNet-18}}  \\
StyleMatch  & \(\checkmark\) & 79.43 & 73.75& 90.04 & 78.40 &80.41 & 78.54 & 74.44 & 89.25 &79.06 &80.32 \\
MultiMatch   & \(\checkmark\)&80.69& 70.44& 90.48& 74.79& 79.10& 79.54& 71.26 &88.00& 73.91 &78.18 \\
CAT (Ours)  & \(\checkmark\)& \textbf{83.04} & \textbf{75.23} & \textbf{93.43} & \textbf{80.10} & \textbf{82.95} &\textbf{82.83} &\textbf{75.50} & \textbf{93.00} & \textbf{76.19} & \textbf{82.71}\\
\bottomrule
\rowcolor{gray!20} \multicolumn{12}{c}{\textbf{ResNet-50}}  \\
StyleMatch  & \(\checkmark\) & 81.72 & 76.19 & 92.58 & 80.04 & 82.45 & 80.18 & 76.58 & 91.09 & 81.42 & 82.96 \\
MultiMatch   & \(\checkmark\) & 83.23 & 72.48 & 92.87 & 77.23 & 81.49 & 81.59 & 73.63 & 90.30 & 76.45 & 80.59 \\
CAT (Ours)  & \(\checkmark\) & \textbf{85.38} & \textbf{77.57} & \textbf{95.77} & \textbf{82.44} & \textbf{85.29} & \textbf{85.17} & \textbf{77.84} & \textbf{95.34} & \textbf{78.53} & \textbf{85.05} \\
\bottomrule
\end{tabular}
\caption{Backbone comparison of ResNet-18 and ResNet-50 in SSDG settings.}\label{table6}
\end{table*}

\begin{figure*}[ht!]
    \centering
    \includegraphics[width=0.8\textwidth]{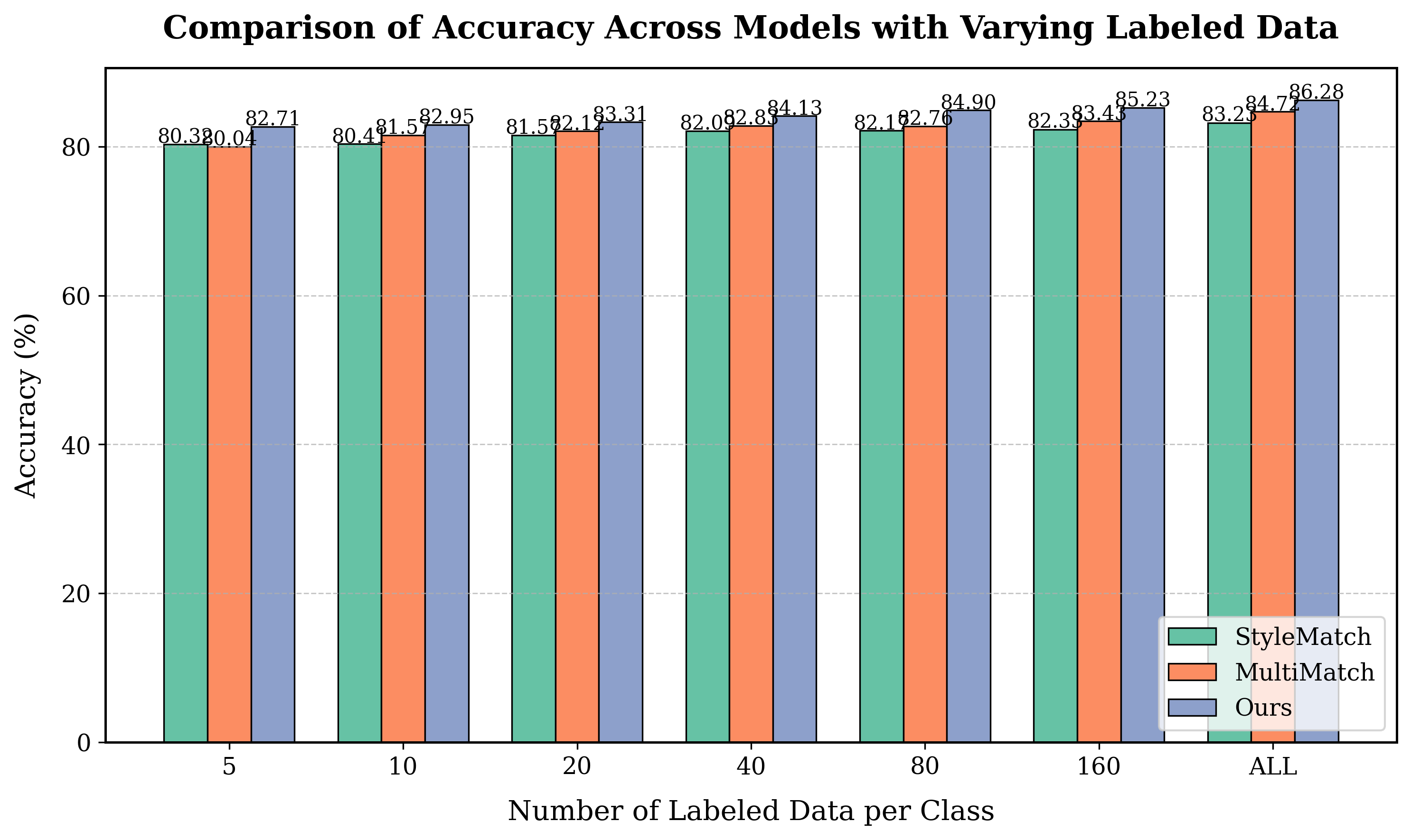}  
    \caption{Comparison between our method with StyleMatch and MultiMatch in different label settings.}
    \label{fig2}  
\end{figure*}

\noindent\textbf{Effect of Different Numbers of Labels.} In Figure \ref{fig2}, we conduct a comparison with different sets of label data to validate the performance of our method. We compare with two SSDG methods, such as StyleMatch, and MultiMatch. In every label set, our method outperforms both StyleMatch and MultiMatch. In all label settings, our method can improve performance by 1.5\% than MultiMatch, which is better 1.5\% better than StyleMatch. Hence, these results demonstrate its effectiveness even in a fully supervised setting.\\

\noindent\textbf{Effect of Different Numbers of Source Domains.} In Table \ref{table7}, we examine the impact of the number of sources ($K$) on the performance of three models—FixMatch, StyleMatch, and CAT (the proposed method)—on the PACS dataset, under two settings of label availability: 10 labels per class and 5 labels per class. The results, reported as accuracy percentages, highlight the influence of $K$ (number of source domains) and the availability of labeled data on the models' performance. The results reveal that increasing the number of sources ($K$) consistently improves accuracy across all models. For instance, FixMatch shows notable improvements as $K$ increases from 1 to 3, but it lags behind StyleMatch and CAT in every configuration. StyleMatch demonstrates better utilization of domain information, consistently outperforming FixMatch across both label regimes. However, CAT significantly surpasses both FixMatch and StyleMatch in all scenarios, indicating its superior capability in leveraging both labeled and unlabeled data for domain generalization. With 10 labels per class, CAT achieves the highest accuracy, with 61.32\% for $K=1$, 78.92\% for $K=2$, and 82.95\% for $K=3$. Even in the low-data regime of 5 labels per class, CAT maintains its dominance, achieving 57.64\% for $K=1$, 74.26\% for $K=2$, and 82.71\% for $K=3$. These results highlight the model’s robustness and scalability, particularly as the number of source domains ($K$) increases. In summary, the findings demonstrate that CAT consistently outperforms FixMatch and StyleMatch, especially as the number of sources grows. Furthermore, it shows remarkable robustness in low-data scenarios, confirming its effectiveness in domain generalization tasks under varying conditions of labeled data availability.\\

\begin{table}[h!]
\centering
\scalebox{0.8}{\begin{tabular}{lcccccc}
\toprule
\multirow{2}{*}{Model} & \multicolumn{3}{c}{10 labels} & \multicolumn{3}{c}{5 labels} \\
\cmidrule(lr){2-4} \cmidrule(lr){5-7}
 & $K=1$ & $K=2$ & $K=3$ & $K=1$ & $K=2$ & $K=3$ \\
\midrule
FixMatch & 53.55 & 71.42 & 77.12 & 49.91 & 68.52 & 74.94 \\
StyleMatch & 57.29 & 74.50 & 80.41 & 52.24 & 71.95 & 80.32 \\
CAT (Ours) & \textbf{61.32} & \textbf{78.92} & \textbf{82.95} & \textbf{57.64} & \textbf{74.26} & \textbf{82.71} \\
\bottomrule
\end{tabular}}
\caption{Impact on the number of sources ($K$) on the PACS dataset with varying label availability: 10 labels per class and 5 labels per class.}
\label{table7}
\end{table}

\section{Conclusion}
In this work, we explore the challenging area of semi-supervised domain generalization (SSDG) to handle domain shifts under a low-data regime. In recent years, SSDG has become a more practical solution for many real-world applications. Hence, we propose \textbf{CAT}, an SSDG method that addresses the limitations of existing approaches by leveraging adaptive thresholding and noisy label refinement techniques to generate reliable pseudo-labels and enhance generalization. By employing both global and local adaptive thresholds, our method ensures improved class diversity and dynamic confidence management in pseudo-label generation. Additionally, the integration of supervised contrastive learning with refined pseudo-labels enables the model to capture domain-invariant representations effectively. Experimental results demonstrate the effectiveness of our method as an SSDG solution. 

\section*{Acknowledgement}

\noindent This research is supported by Hallym University Research Fund, 2024 (HRF-202408-001).

{\small
\bibliographystyle{ieee_fullname}
\bibliography{new}
}

\end{document}